\documentclass[10pt,twocolumn,letterpaper]{article}

\usepackage{cvpr}
\usepackage{times}
\usepackage{epsfig}
\usepackage{graphicx}
\usepackage{amsmath}
\usepackage{amssymb}
\usepackage{color, colortbl}

\usepackage{float}
\usepackage{subfig}

\usepackage[pagebackref=true,breaklinks=true,letterpaper=true,colorlinks,bookmarks=false]{hyperref}

\cvprfinalcopy 

\def\cvprPaperID{560} 

\definecolor{lightgray}{gray}{0.4}

\newcommand{\chg}[1]{{\color{red}#1}}
\renewcommand{\chg}{}

\newcommand{\A}{\mathcal{A}}
\renewcommand{\S}{\mathcal{S}}

\newcommand{\I}{\mathcal{I}}

\newcommand{\D}{\mathcal{D}}

\newcommand{\T}{\mathcal{T}}

\renewcommand{\P}{\mathcal{P}}
\newcommand{\R}{\mathbb{R}}
\newcommand{\F}{\mathcal{F}}

\DeclareMathOperator*{\argminx}{argmin} 
\newcommand{\argmin}[1]{\argminx_{#1}}

\newcommand{\MACE}{\text{MACE}}
\newcommand{\MRE}{\text{MRE}}

\usepackage{xspace}
\usepackage{multirow}

\definecolor{rowCol}{rgb}{0.88,1,1}

\graphicspath{{imgs/}}

\ifcvprfinal\fi

\begin{document}

\title{Unsupervised Deep Single-Image Intrinsic Decomposition\\using Illumination-Varying Image Sequences}

\author{Louis Lettry\\
CVL, ETH Z\"urich\\
{\tt\small lettryl@vision.ee.ethz.ch}
\and
Kenneth Vanhoey\\
CVL, ETH Z\"urich\\
Unity Technologies\\
{\tt\small kenneth@research.kvanhoey.eu}
\and
Luc van Gool\\
CVL, ETH Z\"urich\\
PSI-ESAT, KU Leuven \\
{\tt\small vangool@vision.ee.ethz.ch}
}

\makeatletter
\let\@oldmaketitle\@maketitle
\renewcommand{\@maketitle}
{
\@oldmaketitle

 \centering
  \includegraphics[width=0.75\linewidth]
    {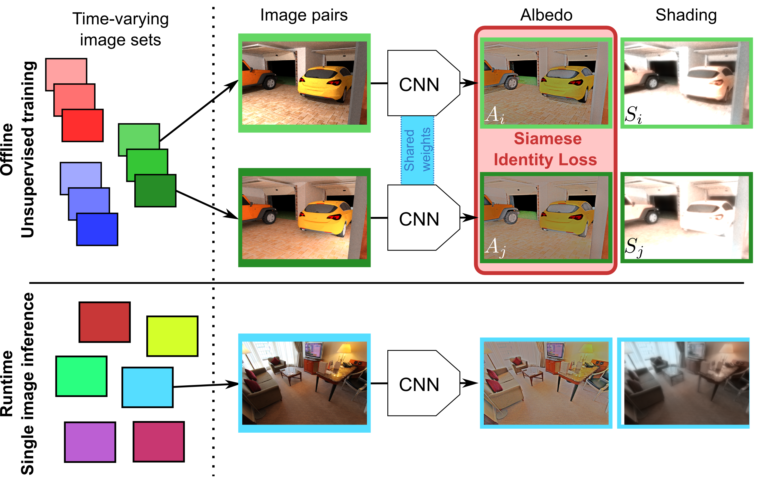}
   \captionof{figure}{An end-to-end convolutional neural network (CNN) trained without ground truth supervision decomposes an input image into its albedo and shading images.  In a pre-computation step (top), a unique CNN is trained by independently processing two images (taken from a time-varying sequence) so that a loss function can be expressed in a siamese manner on both their decompositions comparatively (\ie, red background):  it can learn from observing the changes in the input's shading.  At runtime (bottom), the trained CNN offers decomposition of previously unseen single images.\label{fig:teaser}}
\bigskip
}

\makeatother

\maketitle
\thispagestyle{empty}




\begin{abstract}
Machine learning based Single Image Intrinsic Decomposition (SIID) methods decompose a captured scene into its albedo and shading images by using the knowledge of a large set of known and realistic ground truth decompositions.
Collecting and annotating such a dataset is an approach that cannot scale to sufficient variety and realism.
We free ourselves from this limitation by training on unannotated images.

Our method leverages the observation that two images of the same scene but with different lighting provide useful information on their intrinsic properties: by definition, albedo is invariant to lighting conditions, and cross-combining the estimated albedo of a first image with the estimated shading of a second one should lead back to the second one's input image.
We transcribe this relationship into a siamese training scheme for a deep convolutional neural network that decomposes a single image into albedo and shading. 
The siamese setting allows us to introduce a new loss function including such cross-combinations, and to train solely on (time-lapse) images, discarding the need for any ground truth annotations.

As a result, our method has the good properties of
i) taking advantage of the time-varying information of image sequences in the (pre-computed) training step,
ii) not requiring ground truth data to train on, and
iii) being able to decompose single images of unseen scenes at runtime.
To demonstrate and evaluate our work, we additionally propose a new rendered dataset containing illumination-varying scenes and a set of quantitative metrics to evaluate SIID algorithms.
Despite its unsupervised nature, our results compete with state of the art methods, including supervised and non data-driven methods.

\end{abstract}

\section{Introduction}
Visual acquisition is the result of a complex process in which light travels through a scene and arrives on a sensor.
This measured data is then post-processed (\eg, by a computer or our brain) to form an image representation.
Intrinsic image decomposition is the inverse process in which one tries to recover intrinsic properties of a scene from a single 2D image representation.
Many applications benefit from having access to this disentangled representation to allow for improved scene understanding, feature and shadow detection, stylization, relighting, object insertion and many more~\cite{BonneelSOTA}.
In this paper, we tackle the problem of separating a single image $\I$ into an albedo image $\A$ (\ie, a lighting and acquisition-independent representation of the scene) and its complementary image termed shading\footnote{The term shading refers to light-induced effects, yet we (abusively) also include acquisition-induced effects --~like tone-mapping~-- in this term.} $\S$.

Formally, Single-Image Intrinsic Decomposition (SIID) infers $\A$ and $\S$ from the input image $\I$ such that the pixel-wise product $\I=\A\cdot\S$ is respected~\cite{BT78} (see Fig.~\ref{fig:teaser}).
$\A$ is a color image whose pixels represent the base color or (diffuse) reflectance of what constitutes the acquired scene.
$\S$ is also an image whose pixels represent the alteration of the base color when the local illumination hits the surface and gets reflected towards the viewer.
$\S$ can be colored as well, due to colored illumination (e.g., due to sunset illumination) or inter-reflections on neighboring objects (see Fig.~\ref{fig:model_needs} and~\ref{fig:shadingGT}).
As a result, SIID is severely ill-posed: there are twice more unknowns than knowns in the equation.

To guide an optimization procedure towards a desirable goal for this under-determined problem,
one can define priors based on observation of a simplified Mondrian world (\eg, $\A$ is piecewise constant or $\S$ is smooth~\cite{retinex}),
reduce degrees of freedom (\eg, $\S$ is greyscale~\cite{MITintrinsicImages,zhou2015learning,IIW}) and/or provide additional input data (\eg, image depth~\cite{Chen13}).
On real-world use cases however, automatic SIID remains unsolved~\cite{BonneelSOTA}.
Machine learning offers a way to learn valid priors, rather than impose them.
However, this requires annotated ground truth (GT) data (\ie, dense per-pixel triplets $(\I,\A,\S)$ in our case), which cannot be obtained through a scalable process~\cite{MITintrinsicImages}.
We avoid this constraint by training without GT supervision.

We note that observing time/illumination changes of a static scene provides useful information to guide the learning process.
Indeed, albedo is invariant to changes in illumination.
Hence, two images from the same scene taken at different times should be decomposed into the same albedo and varying shading.
By setting up a siamese training scheme, in which two images get processed in parallel by the same convolutional neural network (CNN), we can express loss functions that encompass both their decompositions.
This way, the network gets optimized based on the relationship of decompositions it produces for pairs of images (see Fig.~\ref{fig:teaser}, top).
As a result, by observing pairs and adding a small regularization, our method learns how to process SIID in an unsupervised manner: it trains solely on (time-lapse) images, discarding the need for any GT annotation.
To our knowledge, it is the first deep learning solution for SIID to be unsupervised thus to avoid training on datasets that are infeasible to annotate or to which one risks severe overfitting.
Moreover, the CNN forms a feed-forward network that can process a single image of a previously unseen scene (see Fig.~\ref{fig:teaser}, bottom): it does not require multiple inputs at runtime.

In the following sections, we build, demonstrate and evaluate our novel method.
More specifically:
\begin{itemize}
 \item In section~\ref{sec:siamese}, we introduce a siamese training procedure that trains a CNN on pairs of images taken from a time-varying scene.
 This allows us to phrase the loss functions that rule the relationship between the estimated decompositions of both images.
 Despite the lack of GT supervision, we obtain results that compete with the state of the art methods.
 \item In section~\ref{sec:results} we propose to complete the set of evaluation metrics used to benchmark SIID algorithms.
 We show that existing numerical benchmarks presented and used separately or jointly in different papers are insufficient to capture all desirable properties of SIID algorithms.
 Therefore, we build a set of metrics --~two of which are new~-- and interpret what each one measures before comparing our results to related work using them.
 \item Additionally, in section~\ref{sec:dataset}, we propose a synthetic dataset that is generated using physically-based rendering.
  It is an extension of the recent SUNCG dataset~\cite{SUNCG}, which we augment for unsupervised SIID by rendering static scenes under varying lighting conditions and tone mappings.
\end{itemize}


\section{Related Work}
\label{sec:rw}
Our interests lie with previous Intrinsic Image Decomposition (IID) methods (section~\ref{sec:rw:ID}), datasets used for training and evaluation of SIID algorithms (section~\ref{sec:rw:datasets}) and evaluation metrics (which is discussed in section~\ref{sec:results}).

\subsection{Intrinsic decomposition methods}
\label{sec:rw:ID}
Single-image IID is the process of decomposing an image $\I$ into a dense albedo map $\A$ and shading map $\S$.
A recent survey by \textit{Bonneel}~\etal~\cite{BonneelSOTA} reviews and compares related work on its usability for typical CG applications.
The problem is severely ambiguous: twice as many unknowns as knowns have to be estimated.
Providing more constraints is thus essential to guide an optimization scheme and reduce degrees of freedom.

\subsubsection{Reducing degrees of freedom}
\paragraph*{Human-devised priors} based on observation (\eg, $\A$ is globally sparse and piecewise constant and $\S$ is smooth~\cite{retinex}) have been used to regularize optimizations.
Many derivatives exist~\cite{ShenReflectance11,Gehler11,GarcesIntrinsic2012,BarronTPAMI2015,ZhaoIntrinsic2014}.
Most generate decent decompositions on Mondrian-like images, but none generalize to the true complexity of photographed everyday scenes.
We believe no human-devised priors can fully capture the complex reality, hence we prefer learning from data as much as possible.

\paragraph*{Time-varying input data.}
Another source for disambiguation of the optimization problem is additional information: \eg, user input~\cite{BPD09,BSTSPP14} or depth information~\cite{Chen13}.
\chg{Methods processing streams of images of varying viewpoints have been developped as~\cite{mekaSIGGRAPH2016} which uses temporal feedback to optimize on newly acquired frames in real-time.}
Weiss' seminal paper~\cite{Weiss01} and derivatives~\cite{MLKS04,LaffontBazin15,Yu16} propose a method that takes multiple images depicting a static scene with temporal variation as input, and outputs a single (constant) albedo image along with a list of (varying) shading images, one per input image.
While these works bear some similarity with ours, there is a fundamental difference in applicability and generalizability.

Their work is based on an iterative algorithm that optimizes a loss function with respect to a full sequence of input images.
Therefore, the full sequence needs to be available at runtime, and the result is valid for this sequence only.
We similarly use time-varying tuples (many pairs of varying scenes in our case) and optimize with respect to the decompositions of all inputs, but only in a precomputation step, \ie, at train time.
At inference time however, our learned model is applicable to \emph{single images} of \emph{unseen scenes}, hence forms a universal end-to-end SIID method.

\paragraph*{Resort to sub-problems.}
Finally, other priors restrict the range of values for $\S$ (\eg, grayscale and/or $\S \leqslant 1$, which disallows specularities).
The overwhelming majority of IID works make such choices~\cite{BonneelSOTA}.
Notably,~\cite{nonlambertianshapenet} defines $\I=\A\D+\S_p$, where a diffuse $\D < 1$ and the specular $\S_p \in\R^{+}$ components are colored.
Similarly, we use the most general and harder variant ($\I=\A\cdot\S$, $\S \in \R^{3+}$).
We motivate this choice in section~\ref{sec:siamese:model}.

\subsubsection{Learning-based solutions}
A nowadays popular and promising trend is to learn the correct priors from data, either explicitely before feeding them to classical optimizations (\eg, leveraging a CRF)~\cite{IIW,zhou2015learning} or implicitely in an end-to-end framework~\cite{DI,nonlambertianshapenet,DARN}.
The largest problem in the task of SIID concerns the source of data to train on.
Since albedo cannot be observed without light, observing GT albedo and shading separately is hardly possible in the real world, and all existing datasets have strong limitations (see Sec.~\ref{sec:rw:datasets}).
Supervised end-to-end learning solutions rely on synthetic datasets~\cite{DI,DARN,nonlambertianshapenet}.
As a result they suffer from overfitting on too small or artificial, biased datasets~\cite{DI}, hindering generalizability to real-world photographs.
\chg{New efforts have been made to alleviate this generalizability problem, as \cite{selfsupervisedintrinsicNIPS17} who proposed a self-supervised approach in two steps: first a rendering network is trained in a supervised manner to estimate a shading given a normal map and a point light, then a decomposing network is trained through the rendering network and a self reconstruction loss to estimate the albedo, normal map and illumination of the input image.}
Others rely on sparse human annotations, followed by applying the classical hand-devised priors~\cite{IIW,zhou2015learning}: the learned model is by definition human-centric thus may be limited in generalizability.
Conversely, we propose the first unsupervised end-to-end deep learning solution that does not rely on any annotation.

\subsection{Datasets for Intrinsic Image Decomposition}
\label{sec:rw:datasets}
\begin{figure}
    \begin{center}
        \subfloat[MIT]{\includegraphics[width=0.3\linewidth,height=20mm]{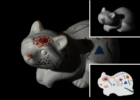}}\ 
        \subfloat[MPI Sintel]{\includegraphics[width=0.3\linewidth,height=20mm]{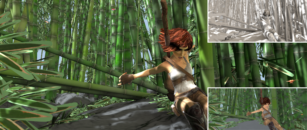}}\ 
        \subfloat[ShapeNet]{\includegraphics[width=0.3\linewidth,height=20mm]{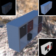}}\\[-3mm]
        \subfloat[IIW/SAW]{\includegraphics[width=0.3\linewidth,height=20mm]{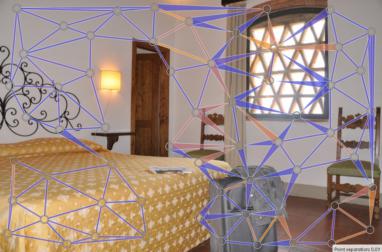}}\ 
        \subfloat[Webcams]{\includegraphics[width=0.3\linewidth,height=20mm]{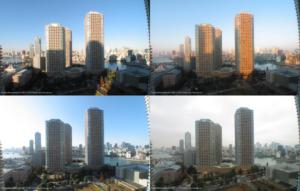}}\ 
        \subfloat[Light Compositing]{\includegraphics[width=0.3\linewidth,height=20mm]{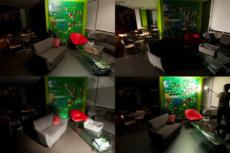}}
    \end{center}
    \caption{One sample per existing dataset. (a-c) is available with dense GT annotations (insets), (d) with sparse relative ones (arrows), and (e-f) \chg{possess no groundtruth.}}
        \label{fig:datasets}
\end{figure}

Datasets for IID allow for two things: training supervised learning-based methods on, and numerical evaluation.
Such a dataset should preferably come with dense GT annotations, \ie, $(\I,\A,\S)$ triplets, but this is expensive to create at best, resulting in a realism versus size/pixel density tradeoff.
We discuss the capabilities and limitations of five datasets illustrated in Fig.~\ref{fig:datasets}(a-d).

\subsubsection{Realistic and Scarce}
\textbf{MIT IID}~\cite{MITintrinsicImages} is the only GT dataset on real-world data.
It contains 20 single-object scenes lit by 10 different illumination conditions.
Dense (\ie, pixelwise) decompositions were defined after a tedious acquisition process involving controlled light and paint-sprayed objects.
Its small size and lack of variety makes it unusable for training convolutional neural networks (CNN) but provides a benchmark for evaluating object decompositions.

\textbf{Intrinsic Images in the Wild}~\cite{IIW} (IIW) introduces a dataset of 5,230 real-life indoor images with sparse reflectance annotations by humans,
who were asked to compare (similar, greater or smaller than) albedo intensity (\ie, grayscale level) of random point pairs in the images.
This is taken as a sparse GT reference to measure the Weighted Human Disagreement Rate (WHDR) of SIID algorithms applied on the IIW images.
Training a dense regression CNN is feasible but the sparse annotations provide insufficient cues (see Section.~\ref{sec:results}).

\textbf{Shading Annotations in the Wild}~\cite{SAW} (SAW) extends and complements IIW~\cite{IIW} with partly dense shading annotations by humans, who were asked to classify pixels as belonging to either smooth shadow areas or non-smooth shadow boundaries.
It is taken as a semi-dense GT reference to measure the SAW quality of SIID algorithms applied on the SAW dataset.

We include all these measures in our panel of metrics measuring various aspects of SIID quality in section~\ref{sec:results}.

\subsubsection{Synthetic and Dense}
CG rendering algorithms allow for approaching photorealistic image quality while accessing and exporting intrinsic layers, like albedo: they offer dense GT by definition.
However, creating a dataset that is completely realistic and covers the visual variety of the real world is impossible, since realistic rendering requires substantial expert human effort and computation time.

\textbf{MPI Sintel}~\cite{SINTEL} contains frames from 48 scenes (along with GT albedo and shading) of the Sintel CG short movie. 
However, it is biased: non-realistic effects (\eg, fluorescent fluids) and harmful modeling tricks (\eg, shadow baking in the albedo) have been used, so training on it hardly generalizes to real-world images~\cite{DI,DARN}.

\textbf{Non-Lambertian ShapeNet}~\cite{nonlambertianshapenet} is closer to photo-realism, but contains only single-objects, just like MIT IID.
25K ShapeNet~\cite{ShapeNet} objects' intrinsic layers were lit by 98 different HDR environment maps and rendered using Mitsuba~\cite{mitsuba}, for a total of 2.4M training images.

\section{Unsupervised siamese training}
\label{sec:siamese}
We wish to avoid the usage of human-devised priors as much as possible to guide a learned solution, which is why we believe data-driven approaches are adequate.
Deep learning brings a renewed interest for solving the task of SIID, but as we have seen in section~\ref{sec:rw:datasets}, supervised learning requires large-scale annotated datasets.
We propose a first solution using unsupervised deep learning leveraging pairs of images in which only the shading has changed.
In section~\ref{sec:dataset} we detail our training data, while we here focus on the model and training goals.

\subsection{Model \& architecture}
\label{sec:siamese:model}
\begin{figure}
    \begin{center}
        \subfloat{\includegraphics[width=0.34\linewidth,height=18mm]{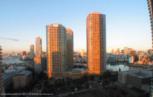}}\ 
        \subfloat{\includegraphics[width=0.34\linewidth,height=18mm]{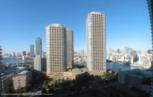}}\ 
        \subfloat{\includegraphics[width=0.30\linewidth,height=18mm]{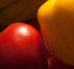}}
    \end{center}
    \caption{Real-life photographs.
    Left: two views of the same scene at different times: the color variation resides in the lighting and acquisition process.
    Right: red light reflected onto the yellow pepper, and a specularity.
    Capturing these effects in $\S$ requires it to be color-valued and without upper bound.
    }
    \label{fig:model_needs}
\end{figure}

Our goal is to build a universal model that decomposes an image $\I$ into an albedo $\A$ and shading $\S$ using an image-to-image regression CNN $\F_c$ with parameters $c$ such that $(\A,\S) = \F_c(\I)$.
For generality, we choose the albedo/shading decomposition following
\begin{equation}
\label{eq:fundamental}
\I = \A\cdot \S\,,
\end{equation}
where $\cdot$ denotes a per-pixel product, $\I^{m \times n \times 3}$, $\A^{m \times n \times 3} \in [0,1]$ and $\S^{m \times n \times 3} \in [0, \infty[$.
$\A$ represents intrinsic colors, which we represent using the usual 3-channel RGB values.
Unlike most related work, we allow $\S$ to be colored and to grow beyond unit value.
This allows to represent natural light phenomena, like colored lighting and bright highlights (see the cityscape or the red reflection on the yellow pepper in Fig.~\ref{fig:model_needs}).
Note that solving this problem is harder than many variants in which shading is grayscale (\ie, single-channel), or disallows specularities (\ie, $\S \in [0, 1]$).
Yet we wish our network to be universal for the SIID problem.
In section~\ref{sec:siamese:losses}, we slightly constrain this increased liberty to guide the training towards a viable solution.

\begin{figure}[t]
    \begin{center}
        \includegraphics[width=0.95\linewidth]{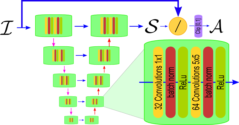}
        \caption{\label{fig:cnn_arch} Our architecture is an autoencoder with skip connections (blue arrows), strided maxpooling (magenta arrows) and upscaling (red arrows).}
        \label{fig:architecture}
    \end{center}
\end{figure}

We choose an architecture in which $\S$ is regressed and $\A$ is deduced by element-wise division\footnote{we empirically observed the same behavior as~\cite{DARN}, regressing S and deducing A produced better results.}, which guarantees consistent results.
It builds upon the latest trends in training image-to-image CNN's and is summarized in Fig.~\ref{fig:cnn_arch}.
Our network is composed of an autoencoder with skip connections at every level.
The data is downsampled (by maxpooling with a stride of size $2$), respectively upsampled (by a bilinear interpolating upscaling operation of size $2$ as well).
Every level of the encoder and decoder is composed of a succession of 2 convolutions-batchnorm-ReLu layers: the first convolution applies a projection of the feature space with $1\times 1$ filters into a $32$-dimensional space, while the second one has $64$ filters of size $5\times 5$.
Finally, we use the element-wise division presented in~\cite{DARN} to enforce consistency of the decomposition.
Note that we add a clipping layer so as to force $\A \in [0, 1]$.
Since division is derivable, both $\A$ and $\S$ can be used in loss functions, allowing for backpropagation of errors on both components simultaneously.
Training has been done with $2$ siamese images (randomly taken from the same image sequence) in mini-batches of size six.
We used the Adam~\cite{adamopti} optimizer with a learning rate exponentially decreasing from $10^{-3}$ to $10^{-5}$ over 30k iterations, taking 22h on an NVidia GTX Titan X.

\subsection{Siamese training losses}
\label{sec:siamese:losses}
Following the assumption of static, time-varying scenes with illumination changes, a natural constraint arises: between images of the same view, only the shading is changing, and albedo is fixed.
We implement this in a siamese training procedure in which we train a network on pairs of images ($\I_i$, $\I_j$) taken from a time-varying image sequence $\T=\{\forall i, \I_i\}$.
For each image $\I_i$ a forward pass generates a decomposition pair $(\A_i,\S_i)$ (cf. Fig.~\ref{fig:teaser}) 
and a joint loss is backpropagated.
We next present the different components of this loss.
Note that while training is siamese (\ie, requires pairs), inference is done on single images (cf. Fig.~\ref{fig:teaser}, bottom).

\paragraph*{Albedo Similarity.}
Our main training target states \emph{the estimated albedo's of any two images of $\T$ should be as close as possible}, in the $L_2$ sense:
\begin{equation}
\label{eq:siam_loss}
 L_a = ||\A_i - \A_j||^2_2, \mbox{ for } i,j \in \T \,.
\end{equation}
Note that this is equivalent (up to a scaling factor) to the formulation stating that \emph{the cross-product of estimated albedo $\A_i$ and shading $\S_j$ should be as close as possible to input image $\I_j$}:
\begin{equation}
\label{eq:siam_loss2}
 L_a = ||\I_i - \A_j\S_i||^2_2, \mbox{ for } i,j \in \T \,,
\end{equation}
which can be obtained by multiplying equation~\eqref{eq:siam_loss} by $\S_i^2$.
Both formulations gave equivalent results in our experiments.

Nevertheless, it still leaves the problem under-determined.
Without regularization, training will inevitably lead to local pitfalls: \eg, all solutions of the form $\A = \varepsilon$, $\S = \I/\varepsilon$, $\forall \varepsilon \in ]0,1]$.
Hence we add a few regularizing terms.

\paragraph*{Shading Chromaticity Smoothness.}
Unlike many SIID algorithms, our shading model is general: it allows for colored shading ($\S$ is tristimulus) including specularities (\ie, $S$ has no upper bound).
This brings a lot of freedom, and our experience showed that without supervision, this can lead to over-colored shading and dull albedo's.

\begin{figure}
    
    \begin{center}
	\subfloat{\rotatebox{90}{\hspace{4mm}Shading}}\ 
	\subfloat{\rotatebox{90}{from SUNCG-II}}\ 
        \subfloat{\includegraphics[width=0.27\linewidth,height=20mm]{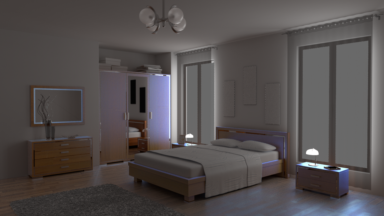}}\ 
        \subfloat{\includegraphics[width=0.27\linewidth,height=20mm]{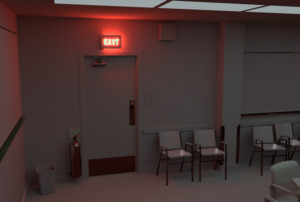}}\ 
        \subfloat{\includegraphics[width=0.27\linewidth,height=20mm]{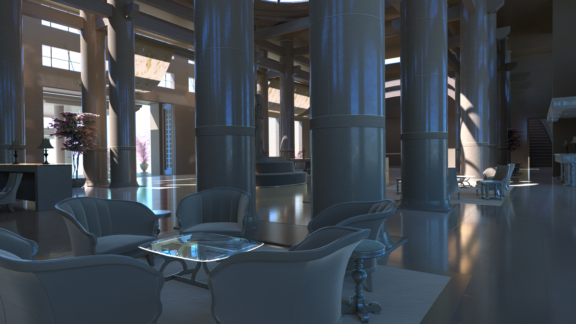}}\\[-3mm]
        \subfloat{\rotatebox{90}{\hspace{4mm}Shading}}\ 
        \subfloat{\rotatebox{90}{\hspace{3mm}chromatic.}}\ 
        \subfloat{\includegraphics[width=0.27\linewidth,height=20mm]{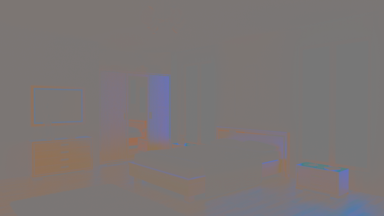}}\ 
        \subfloat{\includegraphics[width=0.27\linewidth,height=20mm]{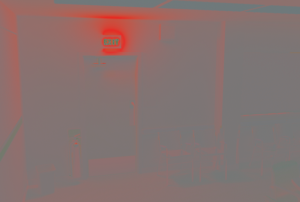}}\ 
        \subfloat{\includegraphics[width=0.27\linewidth,height=20mm]{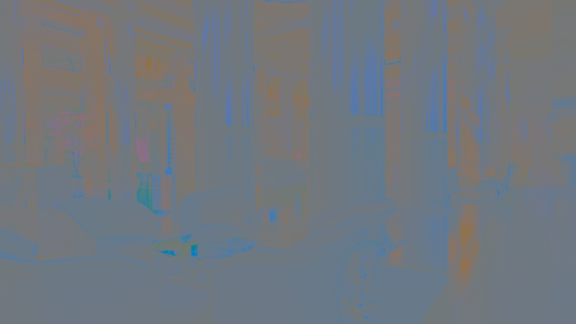}}\\[-2.5mm]
        \subfloat{\rotatebox{90}{\hspace{4mm}Shading}}\ 
        \subfloat{\rotatebox{90}{from \cite{BonneelSOTA}}}\ 
        \subfloat{\includegraphics[width=0.27\linewidth,height=20mm]{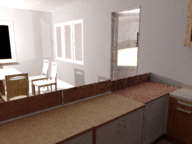}}\ 
        \subfloat{\includegraphics[width=0.27\linewidth,height=20mm]{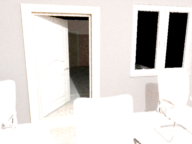}}\ 
        \subfloat{\includegraphics[width=0.27\linewidth,height=20mm]{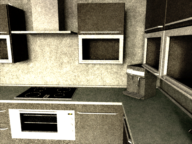}}\\[-3mm]
        \subfloat{\rotatebox{90}{\hspace{4mm}Shading}}\ 
        \subfloat{\rotatebox{90}{\hspace{3mm}chromatic.}}\ 
        \subfloat{\includegraphics[width=0.27\linewidth,height=20mm]{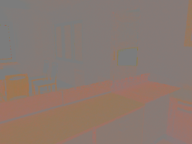}}\ 
        \subfloat{\includegraphics[width=0.27\linewidth,height=20mm]{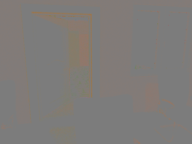}}\ 
        \subfloat{\includegraphics[width=0.27\linewidth,height=20mm]{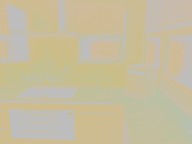}}

     \end{center}

    \caption{Ground-truth shadings obtained with physically-based rendering (odd rows) and its chroma-only image obtained by setting $L=50$ in the CIE-Lab color space (even rows). One can notice that the shadings' chromaticities vary smoothly.}
    \label{fig:shadingGT}
\end{figure}

Shading strongly correlates with geometry (it can contain high frequencies if the geometry does): a shading smoothness loss~\cite{retinex} may thus be undesirable (see Fig.~\ref{fig:shadingGT}, top).
The chromaticity however varies smoothly (Fig.~\ref{fig:shadingGT}, bottom) since light sources tend to be limited in number and colors, and are often distant.
Our formulation ($\S \in [0, +\infty[^3$) allows us to emphasize the regularization on the chromaticity.
To do so, we convert the estimated shading in the CIE-Lab color space and limit the shading gradient of the ab dimension as follows:
\begin{equation}
\label{eq:regul_chroma}
 L_c = \kappa||\nabla\S_{ab}i||^2_2.
\end{equation}
We empirically found that $\kappa$'s ideal range lies in $[10, 100]$, less the loss had basically no impact on the training and more tended to force the shading into a grayscale shading. 
We set it to be equal to $75$, letting the network to favor mainly monochromatic shadings while allowing for colored light changes.

We still keep a small weight on overall shading smoothness in reaction to albedo luminance bleeding into the shading:
\begin{equation}
\label{eq:regul_shading}
 L_c = \lambda||\nabla\S i||^2_2.
\end{equation}
where $\lambda=0.5$ is small so as to let the aforementioned losses take the lead in the optimization.

\paragraph*{Initialization.}
To constrain the remaining degrees of freedom, we make the assumption that most of the albedo's color and texture is well approximated by the actual (temporally-varying) input images.
So we add a loss that encourages albedo to be close input images in the early training stages:
\begin{equation}
\label{eq:regul_init}
 L_i = \mu||\I_j - \A_i||^2_2, \mbox{ for } i\neq j, i,j \in \T \,.
\end{equation}
where $\mu$ decreases linearly from $1$ to $0.01$ during the first 50\% of the training, then remains fixed.
This strongly initializes the model, while loosening it during training in favor of Eqn.~\eqref{eq:siam_loss}.

\begin{figure}
    \begin{center}
        \subfloat[Input $\I$]{\includegraphics[trim={96px 62px 0 10px},clip,width=0.32\linewidth]{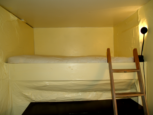}}\ 
        \subfloat[$\A$ w/ cross $(i \neq j)$]{\includegraphics[trim={96px 62px 0 10px},clip,width=0.32\linewidth]{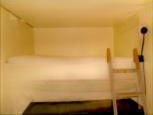}}\
        \subfloat[$\A$ w/o cross $(i=j)$]{\includegraphics[trim={96px 62px 0 10px},clip,width=0.32\linewidth]{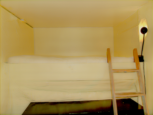}}
    \end{center}
    \caption{Albedo decompositions using the $L_a$ term with and without temporal variation.}
    \label{fig:cross_reg}
\end{figure}

Note that $i \neq j$: we favor proximity between image $\I_j$ at time~$j$ and albedo $\A_i$ at time $i$, hence the name ``Temporal Regularization''.
We also experimented with the simpler variant $i = j$, but this explicitly motivates the network to keep some shading coming from $\I_i$ in its decomposition $\A_i$.
The difference in shading between $\I_j$ and $\I_i$ prevents this undesired effect.
Fig.~\ref{fig:cross_reg} illustrates comparative results: much less shading spills into $\A$ when using $i \neq j$.

\paragraph*{Reconstruction Consistency.}
So as to obtain consistent decompositions (\ie, multiplying $\S$ and $\A$ should produce $\I$ again), following the argument of~\cite{DARN}.
However, because of the clipping layer at the end of the network (see Fig.~\ref{fig:architecture}), and because we do not use any GT supervision as in~\cite{DARN}, consistency could be lost during the optimization.
Hence, we add a loss term to counter this:
\begin{equation}
 L_r = \nu||\I_i - \A_i\S_i||^2_2, \forall i \in \T\,.
\end{equation}
where $\nu=100$ to strongly discourage any deviation from equation~\eqref{eq:fundamental}
In practice, this loss reaches and stays close to 0 after $15\%$ of our train time.


\begin{figure*}
\centering
    \begin{center}
        \subfloat{\includegraphics[width=0.135\linewidth]{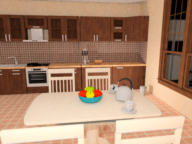}}\ 
        \subfloat{\includegraphics[width=0.135\linewidth]{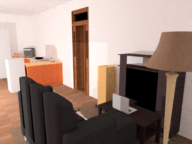}}\ 
        \subfloat{\includegraphics[width=0.135\linewidth]{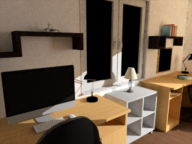}}\ 
        \subfloat{\includegraphics[width=0.135\linewidth]{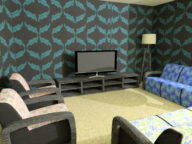}}\ 
        \subfloat{\includegraphics[width=0.135\linewidth]{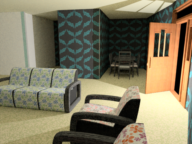}}\ 
        \subfloat{\includegraphics[width=0.135\linewidth]{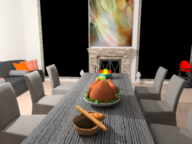}}\ 
        \subfloat{\includegraphics[width=0.135\linewidth]{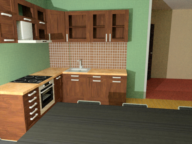}}\\[-3mm]
        
        \subfloat{\includegraphics[width=0.135\linewidth]{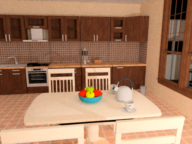}} \ 
        \subfloat{\includegraphics[width=0.135\linewidth]{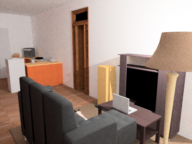}} \ 
        \subfloat{\includegraphics[width=0.135\linewidth]{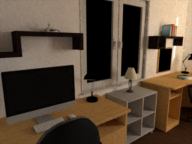}}\ 
        \subfloat{\includegraphics[width=0.135\linewidth]{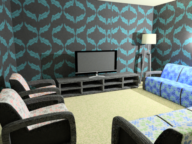}} \ 
        \subfloat{\includegraphics[width=0.135\linewidth]{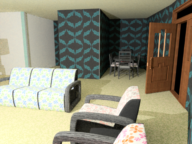}} \ 
        \subfloat{\includegraphics[width=0.135\linewidth]{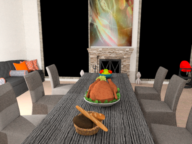}} \ 
        \subfloat{\includegraphics[width=0.135\linewidth]{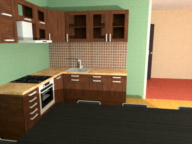}} \\[-3mm] 
        
        \subfloat{\includegraphics[width=0.135\linewidth]{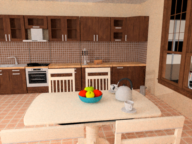}} \ 
        \subfloat{\includegraphics[width=0.135\linewidth]{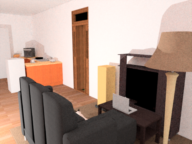}} \ 
        \subfloat{\includegraphics[width=0.135\linewidth]{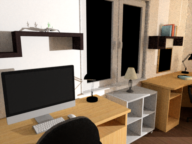}}\ 
        \subfloat{\includegraphics[width=0.135\linewidth]{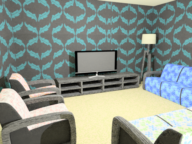}} \ 
        \subfloat{\includegraphics[width=0.135\linewidth]{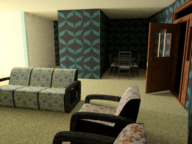}} \ 
        \subfloat{\includegraphics[width=0.135\linewidth]{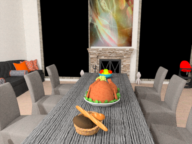}} \ 
        \subfloat{\includegraphics[width=0.135\linewidth]{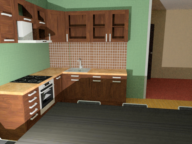}}
    \end{center}
    \caption{SUNCG-II: three variants (varying lighting and tone-mapping) of seven example scenes of our dataset.}
    \label{fig:siam_suncg}
\end{figure*}
\section{Timelapse Datasets}
\label{sec:dataset}
Training CNNs, especially deep ones, requires large-scale GT annotations.
However, both synthetic CG rendering and crowdsourcing human annotations on real images are hardly scalable processes: they are not sufficiently realistic or dense and both require extensive and expert human intervention.
As an alternative, we propose to work on an abundant data source: timelapses.
We define timelapses as a collection of images acquired from a fixed viewpoint of a scene, with time-varying environmental parameters like weather.
Under an assumption of staticity, they observe a constant albedo with different illuminations and acquisition processes (\eg, tone-mapping).

Typical web cameras (see Fig.~\ref{fig:model_needs}, left) form the target training data of our approach, for its ease of acquisition and realism.
However, they often violate the staticity assumption.
For example, the webcamclipart dataset~\cite{webcamclipart} (Fig.~\ref{fig:datasets}(e)) contains 54 webcams that acquired several images per day over a year, and showed that elements may move (including the camera itself), and that weather (\eg, fog, snow) changes the intrinsic albedo of the scene.
We leave sanitization of this data for future work and create a new synthetically rendered dataset in which we have full control over these aspects.

\subsection{SUNCG-II}
We present SUNCG-Intrinsic Images (SUNCG-II), a synthetic dataset that guarantees staticity to train on (cf. Fig.~\ref{fig:siam_suncg}).
It is an extension of the SUNCG dataset, which
is a recent database of modeled apartments and houses introduced in~\cite{SUNCG}.
Geometry, 
aspect 
of each surface, light parameters 
and preset interior viewpoints with full camera calibrations are included.
This data serves as a base to render views (\ie, a fixed scene acquired from a fixed viewpoint and intrinsic camera parameters) with physically-based path tracing using the Mitsuba renderer~\cite{mitsuba} .
The primary objective of SUNCG was to obtain realistic GT interiors for different CV applications such as depth or normal estimation, semantic labeling, or scene completion.

We propose to adapt and augment this dataset to model several shading and image acquisition variants for each static scene and viewpoint.
For each viewpoint, we randomly sample several variants in light sources, and post-process the images with several variants of tone-mapping.
As a result, we created 7,892 views from 817 scenes, multiplied by 5 varying lighting conditions and 5 different tone-mappings, producing a total of 106,609 images (after removing around half of the images because they have too little light, \ie, with mean intensity less than 20).
It is to be noted that SUNCG comes with 45,622 scenes, thus we currently exploited only 1.8\% of the available scenes.

For each image, we have also rendered the corresponding GT albedo and shading maps.
We only use this GT for evaluation, and to compare how far our unsupervised training is off \wrt a supervised variant.
While the dataset we created and will publish is composed of time-varying data with annotated GT, we emphasize that our unsupervised method does not use the GT data.

\chg{
While we think synthetic data should be avoided due its lack of realism, having access to a realistic timelapse rendering framework provides control to every parameters involved in the image generation process (\ie scene, lighting, camera, etc).
This is particularly useful to investigate aspects of the intrinsic decomposition that are often overlooked eventhough they play an important role as these alter the perception of intrinsic properties in a (a priori) uncontrollable manner.
Especially, physical acquisition processes and post processing such as white balancing, tone-mapping, and many others, noticeably modify the acquired image from the original physical scene for perceptual and aesthetic reasons.
}

\subsection*{Technical details}
To determine lighting and tone-mapping variants of each view, we use the following procedure.
First, we remove every transparent object in the scene (\eg, windows, vases) as they incur many rendering artifacts.
Second, we remove any light source, including the environment maps.
Third, we randomly add 1 to 3 point lights in a half cuboid of radius $3\times 1.5 \times 3$ scene units (in camera reference) in front of the camera.
Finally, we render the scene and apply the post-processing tonemap operation~\cite{Reinhardtonamp2002} with parameters $\mbox{\emph{key}}\sim U(0.1,0.6)$ and $\mbox{\emph{burn}} \sim U(0.0,0.2)$.

The photorealistic path-traced images ($\I$) were rendered with 128 samples per pixel.
Albedo maps ($\A$) were rendered by fetching the material's diffuse color/texture information only,
while shading maps ($\S$) were calculated by element-wise division: $\S=\I/\A$.
Validity masks were also produced, discarding infinite depth points and black pixels (\ie, $\A=0$): these pixels are ignored when training.

\subsection{Experimental setting}
\label{sec:dataset:notation}
For comparison and evaluation purposes, we trained several variants in training goals.
We present them here along with the notations we will use in the results section.
``Our'' denotes the standard unsupervised version of this work, trained on SUNCG-II data.
``Our supervised'' is the supervised variant, trained summing two $L_2$ norms (\wrt the GT from SUNCG-II) on $\A$ and $\S$.
``Our IIW'' uses a different dataset (\ie, IIW~\cite{IIW}) and its sparse annotations with augmentations~\cite{zhou2015learning}: we train for optimizing the WHDR score by supervising the training with the annotated sparse pixel relationships.

Finally, we will use the ``light compositing'' (L.C.) dataset~\cite{LightCompositing} consisting of 6 scenes observed from a single viewpoint but different single-flashlight illuminations (cf. Fig.~\ref{fig:datasets}(f)).
It is too small to be used for training, but forms an interesting dataset for comparing the consistency of albedo decompositions.


\begin{table}[t]
    \begin{center}
        \begin{tabular}[c]{|l|c|c|c|}
            \hline
            LMSE & {MIT} & {Bonneel}\\
            \hline
            \hline
            \cite{GarcesIntrinsic2012}  & 8.28 & 5.31 \\
            \cite{zhou2015learning}       & 6.12 & \textbf{1.04} \\
            \cite{DI}                 & 5.92 & 1.38 \\
            \cite{IIW}                    & 5.59 & 1.43 \\
            \hline \hline
            \textbf{Our}                             & \textbf{3.01} & 1.31\\
            \textbf{Our superv.}                     & 3.27 & 1.79\\
            \hline
        \end{tabular}
    \end{center}
    \caption{LMSE ($\times 10^{-2}$) of the state-of-the-art methods \wrt the GT-annotated datasets MIT~\cite{MITintrinsicImages} and Bonneel~\cite{BonneelSOTA}.}
    \label{tab:MSE}
\end{table}
\begin{figure*}[t]
\centering
    \begin{tabular}{cccccc}
		        Input & GT & Ours & \cite{zhou2015learning} & \cite{GarcesIntrinsic2012} & \cite{ZhaoIntrinsic2014} \\
        \multirow{-2}{*}{\includegraphics[width=0.16\linewidth]{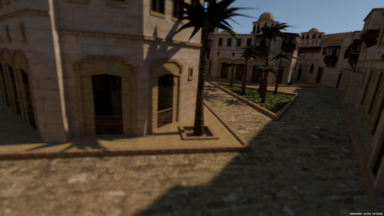}}\hspace{-3mm} & 
        \includegraphics[width=0.16\linewidth]{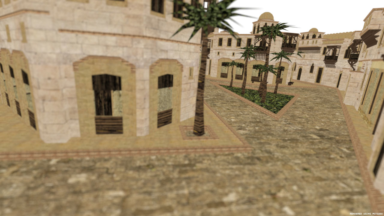}\hspace{-3mm} & 
        \includegraphics[width=0.16\linewidth]{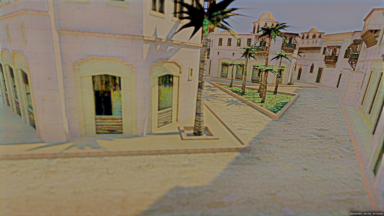}\hspace{-3mm} &
        \includegraphics[width=0.16\linewidth]{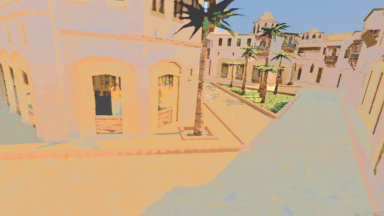}\hspace{-3mm} &
        \includegraphics[width=0.16\linewidth]{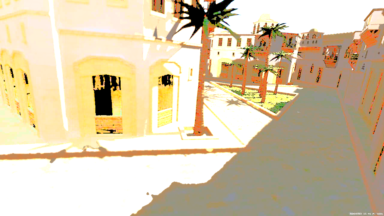}\hspace{-3mm} &
        \includegraphics[width=0.16\linewidth]{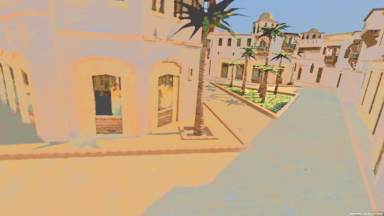}\hspace{-3mm}\\
        &
        \includegraphics[width=0.16\linewidth]{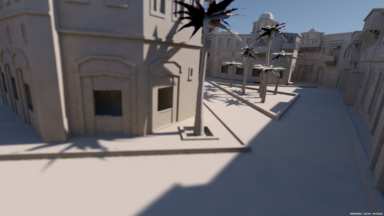}\hspace{-3mm} & 
        \includegraphics[width=0.16\linewidth]{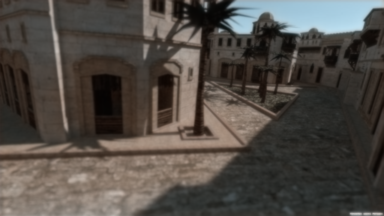}\hspace{-3mm} &
        \includegraphics[width=0.16\linewidth]{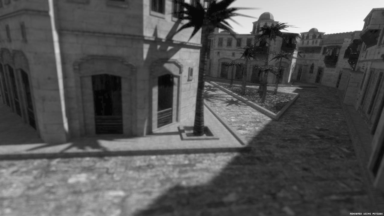}\hspace{-3mm} &
        \includegraphics[width=0.16\linewidth]{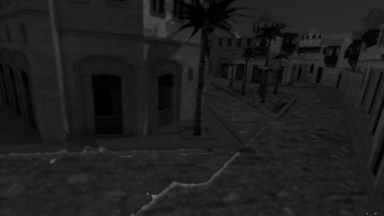}\hspace{-3mm} &
        \includegraphics[width=0.16\linewidth]{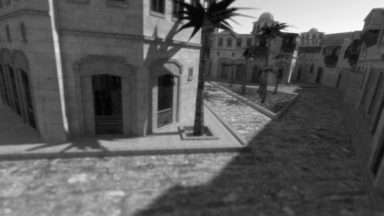}\hspace{-3mm}\\
        
        \multirow{-2}{*}{\includegraphics[width=0.16\linewidth]{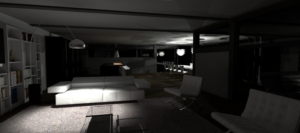}}\hspace{-3mm} & 
        \includegraphics[width=0.16\linewidth]{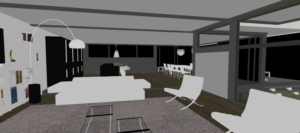}\hspace{-3mm} & 
        \includegraphics[width=0.16\linewidth]{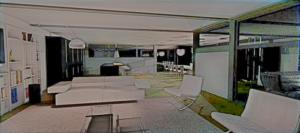}\hspace{-3mm} &
        \includegraphics[width=0.16\linewidth]{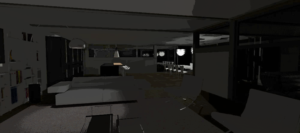}\hspace{-3mm} &
        \includegraphics[width=0.16\linewidth]{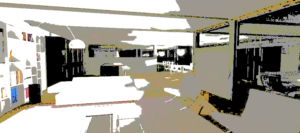}\hspace{-3mm} &
        \includegraphics[width=0.16\linewidth]{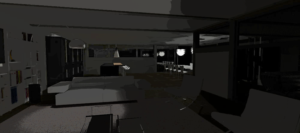}\hspace{-3mm}\\
        &
        \includegraphics[width=0.16\linewidth]{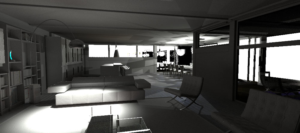}\hspace{-3mm} & 
        \includegraphics[width=0.16\linewidth]{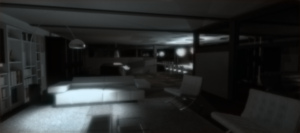}\hspace{-3mm} &
        \includegraphics[width=0.16\linewidth]{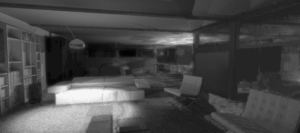}\hspace{-3mm} &
        \includegraphics[width=0.16\linewidth]{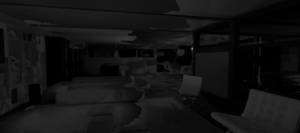}\hspace{-3mm} &
        \includegraphics[width=0.16\linewidth]{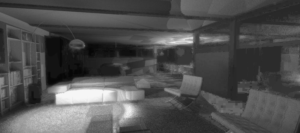}\hspace{-3mm}\\
    \end{tabular}
 \caption{Decompositions on the GT dataset of \emph{Bonneel}~\etal~\cite{BonneelSOTA}.}
 \label{fig:GT}
\end{figure*}

\section{Results}
\label{sec:results}
There are many applications to IID both in computer graphics (\eg, shading-preserving texture editing, shading-less histogram matching, stylization, relighting, object insertion) and computer vision (\eg, scene understanding, robust feature detection for structure from motion, optical flow or segmentation, and shadow detection)~\cite{BonneelSOTA}.
Depending on the target application, one may have different qualitative expectations from a decomposition algorithm: \eg, texture should be preserved in $\A$, or $\A$ and $\S$ should be strictly consistent, \ie, enforce equation~\eqref{eq:fundamental}.

While several metrics~\cite{MITintrinsicImages,IIW,SAW} have been suggested to evaluate IID, it has been observed that none give the full picture~\cite{SAW}.
Therefore, we now assemble and extend a set of metrics that covers many requirements of IID algorithms, \ie:
\begin{itemize}
 \item proximity to dense GT
 \item agreement with human judgments, and
 \item consistency of decomposition.
\end{itemize}
Our argument is that they are all necessary to give the full (or at least a wider) picture: no metric taken alone is sufficient to validate an IID algorithm.

Alongside quantitative measures, we present qualitative results so as to link numbers with visual quality on the recent realistic rendered scenes by Bonneel~\etal~\cite{BonneelSOTA} and on real images from Bell~\etal~\cite{LightCompositing}.
We evaluate and compare using our full set of metrics and show that despite not being supervised, our method competes with state-of-the-art methods on reference-based measures, and surpasses them on consistency of decomposition.
More detailed results are provided in an additional document.

\subsection{Proximity to Dense Ground Truth}
\label{sec:results:MSE}
Ideally, decompositions should lean closely to true decompositions represented by dense GT.
The most widely-used full-reference metric in IID is the Local Mean Squared Error (LMSE)~\cite{MITintrinsicImages,DI,BonneelSOTA}.

First, we evaluate the learning capacity of our unsupervised siamese scheme (denoted ``Our'') by comparing with a fully supervised training equivalent (denoted ``Our supervised'') against SUNCG-II ground truth.
Both use the same random 80/20 scene split on SUNCG-II so as to minimize view similarity between train and test data.
We obtain LMSE errors of $1.16$ and $1.14$, respectively.
Drawing conclusions from this is complex as trainings converge at a different pace and towards different goals.
Nevertheless, it hints that learning from time-varying shading without ground truth is nearly as informative as training with ground truth data.

%
For comparing related work, we measure LMSE \wrt two small datasets having GT annotations in Tab.~\ref{tab:MSE}: the real-world MIT dataset~\cite{MITintrinsicImages} and the (close to) realistic CG dataset of \textit{Bonneel}~\etal~\cite{BonneelSOTA}.
MIT contains simple objects, with a handful albedo's only, while \textit{Bonneel et al.}'s dataset contains higher-frequency albedo details, closer to casual images.
Fig.~\ref{fig:GT} shows qualitative results alongside GT decompositions.
One can notice that the GT shadings are colored, as are ours.

Despite avoiding GT supervision, our method beats classical methods (\eg, ~\cite{GarcesIntrinsic2012}) and leans close to those that use data-driven supervision~\cite{IIW,zhou2015learning,DI} in Tab.~\ref{tab:MSE}.
Note that our supervised variant is not much better on average.
This may hint at a bias towards the training data domain, hindering generalization beyond it.
Our method being unsupervised, it seems to suffer less from this flaw.

\subsection{Agreement with Human Annotations}
\label{sec:results:human}
Large-scale crowdsourced human annotations on real images have been collected in the IIW~\cite{IIW} and SAW~\cite{SAW} papers, respectively.
The corresponding metrics (\ie, WHDR and SAW, see Sec.~\ref{sec:rw:datasets}) measure the alignment of IID results with these annotations.
The SAW metric evaluates the smoothness of the decomposed shading, while the WHDR is a sparse metric comparing the relative relationship between pairs of albedo intensity points (around 65\% equality and 35\% inequality).
Despite being sparse, both metrics are the best there exist up to date, so we quantitatively measure albedo and shading quality using them.
We recalculated all WHDR and SAW values following the fair protocol of~\cite{IIW} taking the best of two runs with and without sRGB to RGB conversion and using the train/test split of~\cite{zhou2015learning}.
Hence our results differ from those presented in~\cite{zhou2015learning,SAW}.

\begin{table}[t]
    \begin{center}
        \begin{tabular}[c]{|l|c|c|c|c|}
            \hline
            & WHDR & \multicolumn{3}{c|}{SAW precision @}   \\
            \hline
            \textbf{Method} & & 50\% & 70\% & 80\% \\
            \hline \hline            
            Constant Albedo                             & 36.5 & 82.7 & 82.2 & 78.7 \\
            \cite{shen11}                    & 24.0 & 90.0 & 78.4 & 70.1 \\ 
            \cite{retinex}                & 23.5 & 93.4 & 87.5 & 77.3 \\
            \cite{GarcesIntrinsic2012}     & 22.6 & 95.8 & 84.7 & 75.4 \\
            \cite{ZhaoIntrinsic2014}         & 23.2 & \textbf{98.3} & 90.2 & 80.4 \\
            \cite{IIW}                       & \textbf{19.2} & 97.8 & 88.9 & 79.1 \\
            \cite{zhou2015learning}          & 20.1 & 97.8 & {92.9} & 80.3 \\
            \cite{DI}                      & 40.7 & 89.2 & 80.9 & 73.9 \\
            \cite{SAW}                     & N/A  & 93.8 & 84.5 & DNC \\

            \hline \hline
            \textbf{Our}                        	& 35.6  & 97.8 & \textbf{95.3} & \textbf{88.3} \\
            \textbf{Our superv.}                    & 36.4  & 91.5 & 75.7 & 64.0 \\
            \hline
        \end{tabular}
    \end{center}
    \caption{WHDR~\cite{IIW} and SAW (precision at given recall values)~\cite{SAW} evaluation.}
	  \label{tab:humanannot}
\end{table}

\begin{figure*}
    \begin{center}
        \subfloat{\includegraphics[width=\linewidth]{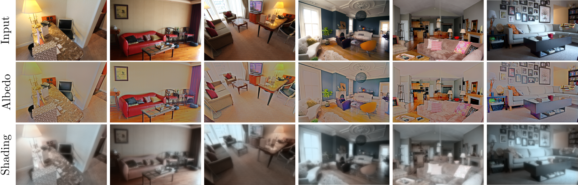}}  
    \end{center}
    \caption{Decompositions using our CNN applied on the IIW dataset.}
    \label{fig:IIW_results}
\end{figure*}

\begin{figure}[t]
\centering
  \setlength\tabcolsep{0.5pt}
    \begin{tabular}{ccccc}
        Input & Ours & \cite{zhou2015learning} & \cite{IIW} \\
        \multirow{-2}{*}{\includegraphics[width=0.24\linewidth]{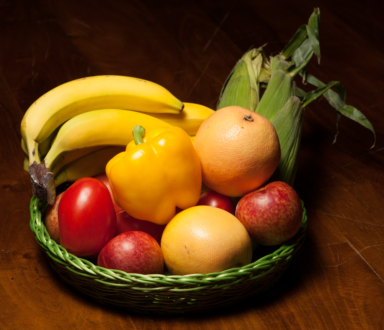}} & 
        \includegraphics[width=0.24\linewidth]{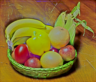} &
        \includegraphics[width=0.24\linewidth]{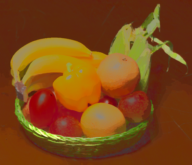}&
        \includegraphics[width=0.24\linewidth]{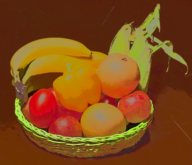} 
        \\[-1mm]
	& 
	\includegraphics[width=0.24\linewidth]{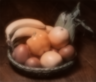} &
	\includegraphics[width=0.24\linewidth]{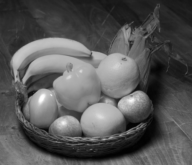} &
	\includegraphics[width=0.24\linewidth]{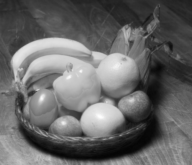} 
	\\[-1mm]
        \multirow{-2}{*}{\includegraphics[width=0.24\linewidth]{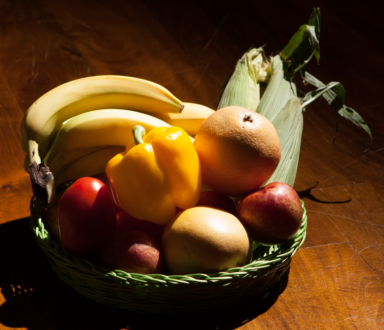}} & 
        \includegraphics[width=0.24\linewidth]{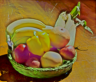}  &
        \includegraphics[width=0.24\linewidth]{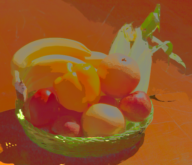}   &
        \includegraphics[width=0.24\linewidth]{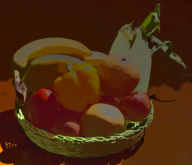}		\\[-1mm]
	& 
	\includegraphics[width=0.24\linewidth]{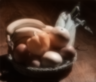}  &
	\includegraphics[width=0.24\linewidth]{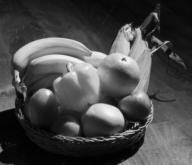}  &
        \includegraphics[width=0.24\linewidth]{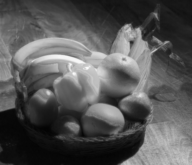}		\\[-1mm]
    \end{tabular}
 \caption{IID on two different lightings per scene on the Light Compositing dataset~\cite{LightCompositing}.}
 \label{fig:consistency}
\end{figure}

WHDR and SAW results can be seen in Tab.~\ref{tab:humanannot}.
Our method lags behind on the WHDR but has an excellent SAW score, especially on high recall values.
With the help of qualitative results in Figures~\ref{fig:GT},~\ref{fig:IIW_results} and~\ref{fig:consistency}, this result can be interpreted as follows.
Our decompositions preserve more texture in the albedo (cf. the floor in Fig.~\ref{fig:GT} and the wooden texture behind the basket in Fig.~\ref{fig:consistency}).
As a result, shadings show less texture, which benefits concordance with the SAW measure.

While this is a good property, it is also harmful for the WHDR: as noted by~\cite{IIW,zhou2015learning}, the WHDR measure does not account for the albedo's high frequencies.
Indeed, it is based on sparse human annotations: only low frequencies are accounted for.
Moreover, chromaticity is also absent since annotations care about \emph{albedo intensity}.
As a result, a method that defines albedo as piece-wise constant (\eg, by pushing texture-induced high frequencies to the shading) or greyscale can have an excellent WHDR.
They tend to misrepresent the shading by including all textures: \eg, ~\cite{zhou2015learning} in Fig.~\ref{fig:consistency}.

We directly observed this in two ways too.
First, when removing shading regularization (equations~\eqref{eq:regul_chroma} and~\eqref{eq:regul_shading}) (which allows colored high frequencies into the shading) our WHDR score improves to $30.0$ while the SAW precision at $80\%$ recall dropped to $82.1$.
Second, we also tried overfitting to IIW annotations (see ``Our IIW'' in Section~\ref{sec:dataset:notation} and Tab.~\ref{tab:humanannot}) using our architecture:
the WHDR improves to $21.1$, but at the cost of dramatic visual results and SAW measure (\eg, $70.6$ at $70\%$ recall).

Nevertheless, despite WHDR being debatable as a metric, we acknowledge there is still too much residue of shading present in our estimated albedo, especially regarding hard shadows.
This forms a limitation of our method and an incentive for future work.

\subsection{Decomposition Consistency}
\label{sec:results:consistency}
Finally, we introduce two new metrics to measure IID requirements that are unattended to in prior work, but are nonetheless important for several applications.

\paragraph*{Reciprocity Error} measures the loss of information occurring during the decomposition by comparing the original image ($\I$) and the reconstructed one ($\A\cdot\S$).
This measure is of predominant importance for applications like image editing and object insertion.
We define the \emph{Mean Reconstruction Error} (MRE) as
\begin{equation}
    \label{eq:MRE}
    \MRE(\I) = \argmin{\alpha} \sum||\I - \alpha\A\cdot\S||
    \,.
\end{equation}
For fairness of comparison with other methods who export 8-bit quantized and rescaled results in image file formats, we proceed similarly.
Moreover, we optimize for the rescaling parameter $\alpha$ to virtually undo potential scalings needed for the 8-bit range quantization or for visualization.
Tiny errors remain due to quantization: when training done and converged, our method should have zero MRE.

\paragraph*{Temporal Inconsistency} measures how much the albedo decompositions of a set of images capturing the same static scene under different (lighting) conditions differ.
This is important for robust shadow detection, relighting and any illumination varying (\ie, video) application.
Let $\T^{\A}=\{\forall i, \A_i\}$ be a set of albedo decompositions of such an image sequence $\T$.
We define the \emph{Mean Albedo Consistency Error} as
\begin{equation}
\label{eq:MACE}
  \MACE(\T^{\A})
  =
  \frac{1}{3 \P |\T^{\A}|^2}
  \sum_{i,j \in \T^{\A}}
  \sum_{c}
  |\A_i^c - \A_j^c|
  \,,
\end{equation}
where $\A_i$ is the albedo decomposition of image $\T_i$, and the sum is normalized by the number of pixels per image $\P$, the number of ordered pairs $(i,j)$, which equals $|\T^{\A}|^2$. 
$c$ runs over colors channels.

Additionally, we define $\MACE_t$: a relaxation of $\MACE$ that does not consider dark pixels that have intensity values below a threshold $t$, and normalizes accordingly.
That is because dark pixels in an input image give little information on what the valid decomposition should be, and algorithms typically have to guess such pixel colors by extrapolatation.
Hence, $\MACE_0$ evaluates extrapolation capabilities, but we also use another value ($t=10$) to assess on the more feasible pixels only.
For $t>0$, we exclude from the calculation any pair $(\A_i, \A_j)$ whose non-dark pixels have small overlap area, \ie, less than $20\%$ of the full image.
\begin{table}[t]
  \setlength\tabcolsep{1.7pt}
    \begin{center}
        \begin{tabular}[c]{|l|r|r|r|r|r|r|}
            \hline
            & \multicolumn{3}{c|}{MRE} & \multicolumn{3}{c|}{$\MACE_{10}$}   \\
            \hline
            \textbf{Method} & \multicolumn{1}{c|}{MIT} & \multicolumn{1}{c|}{L.C.} & \multicolumn{1}{c|}{SAW} & \multicolumn{1}{c|}{MIT} & \multicolumn{1}{c|}{L.C.} &
	    \multicolumn{1}{c|}{Webc.} \\
            \hline \hline
            
            \cite{GarcesIntrinsic2012}     & 9.07 & 15.90 & 24.86 & {9.62} & DNC & 26.37 \\ \relax
            \cite{IIW}                       & 2.28 & 2.84 & 1.53 & 35.85 & 40.87 & 27.76 \\ \relax
            \cite{zhou2015learning}          & 1.87 & 2.32 & 1.57 & 29.51 & 40.23 & 24.21 \\ \relax
            \cite{DI}                      & 9.32 & 11.71 & 18.42 & 30.61 & 30.31 & 25.26\\

            \hline \hline
            \textbf{Our}                      & \textbf{0.12} & \textbf{0.28} & \textbf{0.34} & \textbf{8.59} & \textbf{17.54} & \textbf{11.51} \\
            \textbf{Our supervised}                      & 0.36 & 2.97 & 0.68 & 16.47 & 35.33 & 35.62 \\
            \hline
        \end{tabular}
    \end{center}
    \caption{MRE and MACE metrics (\ie, mean pixel deviation in the range $[0, 255]$). \cite{GarcesIntrinsic2012} did not converge (DNC) on most images of the L.C. sequences.}
	  \label{tab:consistency}
\end{table}

Note that the MPRE~\cite{zhou2015learning} is similar, but defined on products of estimated $\A$ and $\S$ across temporal variation in a sequence.
While elegant, we believe it is not adequate because albedo errors are weighed by shading intensity. 
This attenuates errors in underexposed areas (which we think is legitimate), but emphasizes those made in saturated areas, which is arguable, especially with models where $\S \in [0, +\inf[$.

\paragraph*{Result Analysis.}
Tab.~\ref{tab:consistency} shows consistency results over two time-varying datasets, and the SAW images.
Our method is lossless ($\MRE \approx 0$, only tiny quantization errors remain) and best preserves temporal consistency (see Fig.~\ref{fig:consistency}) across the datasets observed. 
Surprisingly, it is more temporally consistent even on the narrow set of albedos of the MIT dataset.
We believe this is because most methods produce an albedo whose average intensity is close to the input image, and this temporally varies a lot on the MIT and Light Compositing datasets.
Conversely, our method generates consistent intensities, more independent from the lightness.
Zhou~\etal~\cite{zhou2015learning} performs similarly well with a small MRE, but lacks temporal consistency.
While Garces~\etal~\cite{GarcesIntrinsic2012} performs well when its assumptions (\eg, piecewise constant albedo) are satisfied (\ie, on the MIT dataset),
it does not generalize well to real-life situations: it is the lossiest method and does not produce temporally consistent results on the real-life cluttered scenes.

\subsection{Summary}
\begin{figure}
    \centering
    \includegraphics[width=0.725\linewidth,trim={0.4cm 0.4cm 0.125cm 0.4cm},clip]{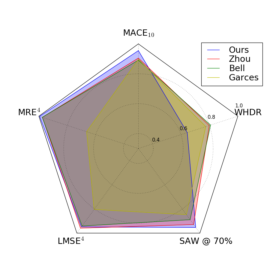}
    \caption{Performance on our five metrics. \chg{The MRE and LMSE are displayed with a power of 4 in order to help visualization.}
  \label{fig:spiderchart}}
\end{figure}
In an effort to gather and concisely present the various properties of the different approaches, we propose to plot them on a spider chart as in Fig.~\ref{fig:spiderchart}.
It visually pinpoints strengths and weaknesses for each method and, from an application point of view, it makes selecting which algorithm suits one's needs easier.
We made it by converting all the metrics to a percentage between 0 and 1, where 1 means no mistake or error (\ie, perfect score).
\chg{The LMSE and MRE have been raised to a power of 4 to improve the visualization by emphasizing the differences.}

Our proposed metrics fill an important gap by measuring aspects of decomposition unmeasured before, and explain how lossy methods such as~\cite{GarcesIntrinsic2012} can perform well on the WHDR or SAW metrics: neither of them penalizes the loss of high frequencies.
Yet, consistency metrics should always be evaluated in the context of reference-based ones or visual results, since trivial solutions (\eg, $\A = \epsilon, \S = \frac{\I}{\epsilon}$) minimize the consistency metrics.

\section{Conclusion \& Limitations}
\label{sec:conclu}
We presented an unsupervised deep learning solution to the single-image intrinsic decomposition problem, which is a first to the best of our knowledge.
It avoids training on datasets that are infeasible to annotate or to which one risks severe overfitting.
In the pre-computation training step, our methodology takes advantage of the relationship between pairs of images of the same scene lit differently.
A CNN gets trained by comparing the result of decomposing two images in parallel, and back-propagating the error to optimize its weights.
Our new loss functions encompass cross-combinations of albedo and shading estimated from image pairs, so it can learn from seeing visual variation with lighting changes.

This allowed us to train the CNN without any kind of GT annotation on our new SUNCG-II dataset.
One of our goals was to eliminate human-designed priors from the optimization goals, as we believe they cannot be general enough to solve the full problem.
Analyzing image by pairs helped for this task, but there are still many degrees of freedom left, as the general problem we try to solve is severely underdetermined.
Hence we resorted to a human-based prior to regularize the optimization (\ie, equation~\eqref{eq:regul_chroma}, and equation~\eqref{eq:regul_init} to a lesser extent).
We see it as a long-term goal to get completely rid of it.

At runtime, the resulting CNN gets deployed on standalone images unseen before, and provides an albedo-shading decomposition.
We evaluate on several SIID metrics, including newly proposed ones, so as to give a large evaluation panel users can choose from depending on their target application.
In general, results compete qualitatively and quantitatively with the state of the art methods, including those that require supervision.
Unlike many methods, our results give high consistency guarantees, which makes our solution a good choice for applications that require consistency on large datasets, \eg, robust feature detection for 3D reconstruction.
Like many methods (see Fig.~\ref{fig:GT}), hard shadows remain difficult to deal with.

We believe the metrics set we assembled brings a broader view on the strengths and weaknesses of each SIID method by evaluating several different or even orthogonal aspects of SIID quality (see Fig.~\ref{fig:spiderchart}).
While defining an overall best method is tricky, this allows the end user to make a more informed choice on what method fits a target application best.
However, linking our 5 quantitative axes with specific applications remains to be thoroughly studied. We leave this as an interesting application-oriented future work.

\chg{Finally, please note that a concurrent work~\cite{BigTimeLi18} proposes a work similar to ours. It also uses timelapses to train CNNs for the task of intrinsic decomposition in an unsupervised manner. They focused on designing a spatio-temporal smoothness that can be efficiently applied to an arbitrary number of input images of the same scene and proposed a sanitized set of timelapses to use for training.}
We believe both our contributions open the path for many exciting works on SIID using unsupervised deep learning.

\paragraph*{Acknowledgments}
The authors acknowledge financial support of the SNF grant ``Wildtrack`` (CRSII2 147693/1) and hardware donation from NVIDIA.

{\small
\bibliographystyle{ieee}
\bibliography{egbib}
}

\end{document}